\documentclass{article}

\usepackage{arxiv}

\usepackage[utf8]{inputenc} % allow utf-8 input
\usepackage[T1]{fontenc}
\usepackage{hyperref}       % hyperlinks
\usepackage{url}            % simple URL typesetting
\usepackage{booktabs}       % professional-quality tables
\usepackage{multirow}       % multi-row table cells
\usepackage{amsmath}
\usepackage{amsfonts}       % blackboard math symbols
\usepackage{nicefrac}       % compact symbols for 1/2, etc.
\usepackage{microtype}      % microtypography
\usepackage{graphicx}
\usepackage{xcolor}
\usepackage{colortbl}
\usepackage{tabularx}
\usepackage{makecell}
\usepackage{wrapfig}
\usepackage{float}
\usepackage{caption}
\usepackage{enumitem}
\usepackage{pifont}
\usepackage{placeins}
\definecolor{vermeerblue}{HTML}{23496D}
\definecolor{vermeermadder}{HTML}{A83E35}
\definecolor{vermeerceladon}{HTML}{1B473C}
\usepackage{array}
\definecolor{ciColor}{HTML}{5A626A}
\definecolor{refgray}{HTML}{F2EFE9}
\definecolor{pendingColor}{HTML}{6B7280}
\definecolor{verdictGreen}{HTML}{0F766E}
\definecolor{verdictAmber}{HTML}{B45309}
\definecolor{hdrbg}{HTML}{F1F5F9}
\definecolor{oursrow}{HTML}{EAF6F4}
\definecolor{oursrowdeep}{HTML}{CBECE4}
\definecolor{ourscell}{HTML}{B8E6DC}
\definecolor{ourstext}{HTML}{0B4F3F}
\definecolor{gaincol}{HTML}{0D7A5F}
\definecolor{badcell}{HTML}{FDE8E8}
\definecolor{badtext}{HTML}{B91C1C}
\definecolor{warnbg}{HTML}{FEF3C7}
\definecolor{warntext}{HTML}{92400E}
\definecolor{refmute}{HTML}{3B495E}
\definecolor{neutralbg}{HTML}{F8FAFC}

\newcommand{\rowstrut}{\rule[-0.95ex]{0pt}{2.85ex}}

\newcommand{\cmarkyes}{\textcolor{ourstext}{\ding{51}}}
\newcommand{\cmarkno}{\textcolor{badtext}{\ding{55}}}
\newcommand{\cmarkpart}{\textcolor{warntext}{$\boldsymbol{\triangle}$}}
\newcommand{\octotitle}{\smash{\raisebox{-0.40\height}{\includegraphics[height=2.65em]{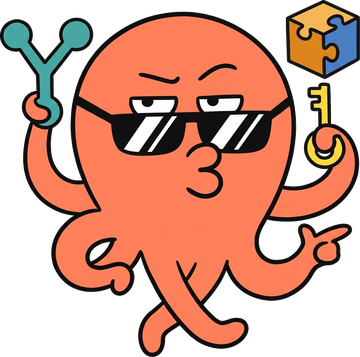}}}\hspace{8pt}}

\newcommand{\octodense}[1]{\rule[-1.50ex]{0pt}{4.15ex}\raisebox{-0.30\height}{\includegraphics[height=1.38em]{figures/#1.png}}\hspace{4pt}}

\newcolumntype{L}[1]{>{\raggedright\arraybackslash}p{#1}}
\newcolumntype{C}[1]{>{\centering\arraybackslash}p{#1}}
\newcolumntype{K}[1]{>{\raggedright\arraybackslash}m{#1}}
\newcolumntype{M}[1]{>{\centering\arraybackslash}m{#1}}
\newcolumntype{Y}{>{\raggedright\arraybackslash}X}
\setlist[itemize]{leftmargin=1.45em, rightmargin=0pt, labelsep=0.45em, itemsep=1.6pt, topsep=2.2pt, parsep=0.3pt}
\setlist[enumerate]{leftmargin=1.55em, rightmargin=0pt, labelsep=0.45em, itemsep=1.6pt, topsep=2.2pt, parsep=0.3pt}

\usepackage[numbers,sort&compress]{natbib}
\usepackage[capitalise,noabbrev]{cleveref}       % smart cross-referencing (load after hyperref)
\usepackage{pgfplots}
\pgfplotsset{compat=1.18}
\pgfplotsset{
  octoaxis/.style={
    scale only axis,
    axis lines=left,
    axis line style={ciColor,line width=0.5pt,-},
    tick style={ciColor,line width=0.4pt},
    tick align=outside,
    major tick length=2.5pt,
    ymajorgrids=true,
    grid style={hdrbg,line width=0.5pt},
    every tick label/.append style={font=\sffamily\scriptsize,color=refmute},
    label style={font=\sffamily\scriptsize,color=refmute},
    title style={font=\sffamily\footnotesize\bfseries,color=refmute,yshift=-3pt},
    clip=false,
  },
  octoline/.style={line width=1.05pt,mark size=1.35pt,solid},
}
\tikzset{octolabel/.style={font=\sffamily\scriptsize,inner sep=1.5pt}}

\renewcommand{\headeright}{}
\renewcommand{\undertitle}{}
\renewcommand{\shorttitle}{Octopus: One Brain, Many Branches}
\title{\mbox{\hspace{4pt}\octotitle\parbox[c]{0.81\textwidth}{\centering\fontsize{15.0}{19.0}\selectfont\scshape Octopus: One Brain, Many Branches---\\Learning Multi-Agent Collaboration through Shared KV Memory}\hspace{24pt}}}

\author{
  \begin{tabular}{@{}C{0.31\textwidth}@{\hspace{0.02\textwidth}}C{0.31\textwidth}@{\hspace{0.02\textwidth}}C{0.31\textwidth}@{}}
    {\normalsize\bfseries Tianming Sha$^{*}$} & {\normalsize\bfseries Junnan Li$^{*}$} & {\normalsize\bfseries Siqi Zhao} \\[2.5pt]
    {\normalfont\small Stony Brook University} & {\normalfont\small University of Wisconsin--Madison} & {\normalfont\small University of Minnesota} \\[1.5pt]
    {\normalfont\footnotesize\texttt{tianming.sha@stonybrook.edu}} & {\normalfont\footnotesize\texttt{junnanl@cs.wisc.edu}} & {\normalfont\footnotesize\texttt{zhao2052@umn.edu}} \\[3pt]
    \multicolumn{3}{c}{\normalfont\footnotesize $^{*}$Equal contribution}
  \end{tabular}
}

\hypersetup{
  pdftitle={Octopus: One Brain, Many Branches -- Learning Multi-Agent Collaboration through Shared KV Memory},
  pdfauthor={Tianming Sha, Junnan Li, Siqi Zhao},
  colorlinks=true,
  linkcolor=vermeermadder,
  citecolor=vermeerblue,
  urlcolor=vermeerblue,
}

\usepackage{xcolor}
\usepackage{pgfplots}
\pgfplotsset{compat=1.18}
\definecolor{oTeal}{HTML}{0D7A5F}    % Octopus
\definecolor{oPurple}{HTML}{6A4CA5}  % sibling reads / mounted KV
\definecolor{oBlue}{HTML}{23496D}    % single-agent and linear baselines; root thread
\definecolor{oGray}{HTML}{8A8F96}    % secondary baseline (always dashed), annotations
\definecolor{oInk}{HTML}{2B2F36}     % axes and tick labels
\definecolor{oGrid}{HTML}{EDEEEF}
\pgfplotsset{
  figaxis/.style={
    scale only axis,
    axis lines=left,
    axis line style={oInk,line width=0.5pt,-},
    tick style={oInk,line width=0.4pt},
    tick align=outside,
    major tick length=2.5pt,
    ymajorgrids=true,
    grid style={oGrid,line width=0.4pt},
    every tick label/.append style={font=\fontfamily{phv}\selectfont\scriptsize,color=oInk,/pgf/number format/assume math mode=true},
    label style={font=\fontfamily{phv}\selectfont\scriptsize,color=oInk},
    title style={font=\fontfamily{phv}\selectfont\footnotesize,color=oInk,yshift=-3pt},
    clip=false,
  },
  figline/.style={line width=1.05pt,mark size=1.35pt,solid},
  figlinealt/.style={line width=1.05pt,mark size=1.35pt,dashed,mark options={solid}},
}
\tikzset{figlabel/.style={font=\fontfamily{phv}\selectfont\scriptsize,inner sep=1.5pt}}

\newcommand{\safefig}[2][]{\ifnum\pdffilesize{#2}0>0 \includegraphics[#1]{#2}\else\fbox{\parbox{0.6\textwidth}{\centering\small Figure pending upload: \detokenize{#2}}}\fi}
\begin{document}
\maketitle\suppressfloats[t]
\vspace{-8pt}
{\small
\begin{abstract}
A long task asks a model to manage its working memory: what to hand off, what to forget, and what to consult. Current agents make these decisions for the model, with rules that compact its context and hierarchies in which each agent sees only what it was handed. We present \textbf{Octopus}, a runtime and a training method that let the model make them itself. Octopus runs one model as a tree of branches over shared KV memory and gives it three operations: a fork hands work to a branch that starts from its parent's state, a rewind forgets a dead end, and a read consults another branch by mounting its memory into attention layers only. On hybrid recurrent-attention models this keeps the reader on its own task, where pasting the same work as text sends most of its following turns astray. DAG-GRPO trains the use of the three operations from outcome rewards alone, treating the whole tree as one sample. On BrowseComp-Plus, a $4$B model trained this way reaches $48.3\%$, the accuracy of OpenAI o3, and outperforms a single agent trained the same way with a window four times as long. It learns all three operations without a reward that names any of them. A $27$B model reaches $64.9\%$ with the same operations and no training.
\end{abstract}
}

\vspace{3pt}
\noindent\centerline{\footnotesize\textbf{Code \& Runtime:}~\href{https://github.com/shatianming5/octopus}{\texttt{https://github.com/shatianming5/octopus}}}

\begin{figure}[H]
\centering
\safefig[width=1.0\textwidth]{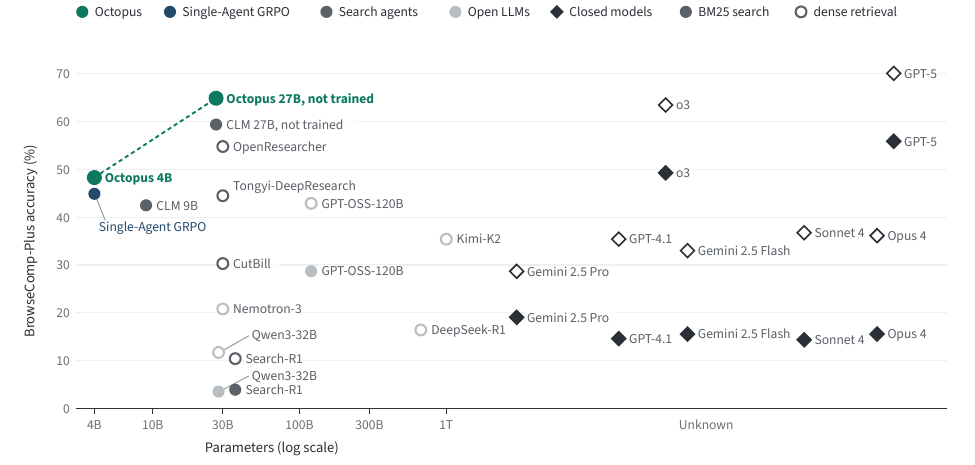}
\caption{Octopus among published systems on BrowseComp-Plus, by model size. Filled markers use BM25 search, as we do, and hollow markers dense retrieval; a system with both results appears twice. Closed models (diamonds), whose sizes are not public, are placed at Unknown in order of release. The dashed line joins Octopus at $4$B, trained, and at $27$B, not trained.}
\label{fig:pareto}
\end{figure}

\section{Introduction}
\label{sec:intro}

\begin{figure}[t]
\centering
\safefig[width=1.0\textwidth]{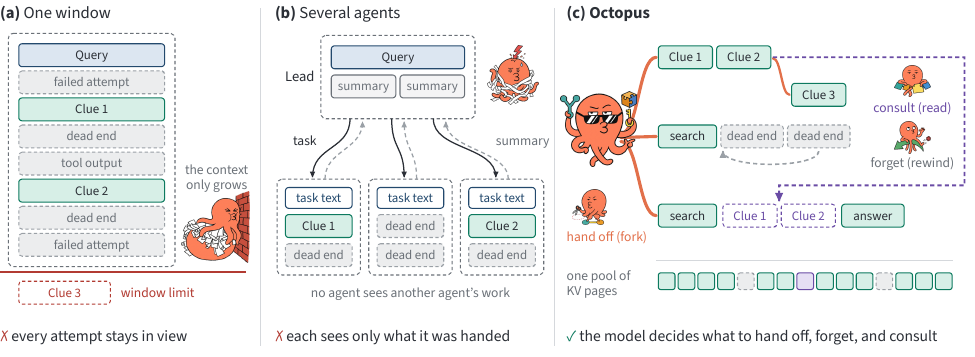}
\caption{Three ways to run a long task: in one window, with several agents that exchange text, and with Octopus, where one model runs a tree of branches over shared memory.}
\label{fig:overview}
\end{figure}

A long task asks a model to manage its working memory: what to hand off, what to forget, and what to consult. A context window lets it do none of these. The window only grows, so two lines of work cannot proceed at once, a failed attempt stays in view, and nothing done elsewhere can be looked up (\Cref{fig:overview}a).

The common way around the window is to use several agents: a lead splits the task, sub-agents work on the parts in windows of their own, and their results come back as messages~\citep{xu2026wideseekr1,hong2025mgrpo} (\Cref{fig:overview}b). This removes the limit of one window and puts a new problem in its place. Each agent sees only what it was handed. What one agent has learned reaches another as a summary, if it reaches it at all, so agents repeat each other's work and act on different pictures of the task. And the hierarchy is fixed before the task begins: sub-agents report to the lead and do not see each other's work. The model works inside these decisions about its memory and does not make them. We give it the operations and ask whether it can learn, from the outcome of the task alone, when to use them.

Learning this requires that handing off, forgetting, and consulting be cheap enough to try often and safe to act on. Two of them can be. A model's working memory is the state it has already computed over its context. Handing off work then means letting a new branch start from its parent's state, and forgetting means dropping the state that a failed attempt wrote. The third is the hard one. A branch that consults another has to read its work, and the usual way is to paste that work in as text. A full transcript rarely fits in the reader's window. A summary fits, but it keeps only what its writer thought to include. And whatever is pasted now sits in the reader's own context, where a long and mixed history already makes a model lose its thread~\citep{liu2024lost}. On hybrid recurrent-attention models we find this at its worst: more than half of the reader's following turns abandon its own task and continue the other branch's. So the question we must answer first is:

\begin{center}
\emph{How can one branch read another branch's work without losing its own task?}
\end{center}

This last failure points to its own fix. A hybrid model keeps its working memory in two forms~\citep{afendulev2026attention}. Its recurrent layers hold a compact state that shapes how the model continues. Its attention layers hold keys and values, the KV cache, which are consulted only when a query asks for them. Pasted text passes through both. We find that in the recurrent layers it overwrites the reader's task, while in the attention layers it is only more to look up. So the reader needs the other branch's keys and values in its attention layers, and nothing in its recurrent layers. Mounted there directly, they also cost no window and need no summary.

Octopus is built on this separation (\Cref{fig:overview}c). One model runs a task as a tree of branches over a shared pool of KV pages, and the three decisions become three operations on that pool. A fork hands off work: the new branch shares its parent's pages and copies its recurrent state. A rewind forgets: it drops the pages of a dead end and restores the state from before it. A read consults: it mounts another branch's pages into the reader's attention layers alone, so the recurrent layers stay on the reader's task. A read uses none of the reader's window, writes no new pages, and keeps off-task turns at $2.7\%$.

The operations are only useful to a model that knows when to use them, and a pretrained model does not. Standard RL training does not apply either, because an episode is now a tree with shared prefixes, removed segments, and edges through which one branch reads another's memory. We introduce DAG-GRPO, which trains the whole tree in one pass from a binary outcome reward and lets gradients flow through the reads.

We study this in deep search, where one question splits into many leads and most of them fail. From the outcome reward alone, a small model learns to hand off work on its first turn, to drop its dead ends, to use a new branch as a fresh window, and to consult the other branches. It reaches the accuracy of OpenAI o3 on BrowseComp-Plus (\Cref{fig:pareto}) and outperforms a single agent trained the same way with a far longer window. A larger model can use the same operations from a prompt, with no training at all.

Our contributions are:
\begin{itemize}
  \item \textbf{Memory operations a model can control.} We let one model run a task as a tree of branches over shared KV memory, with operations to hand off work, forget dead ends, and consult other branches, none of which writes a new page (\Cref{sec:method}).
  \item \textbf{Reading without losing the task.} We show that sharing work as text derails hybrid models, trace the failure to their recurrent layers, and remove it by mounting another branch's memory into attention layers only (\Cref{sec:method-cursor-mount}).
  \item \textbf{Learning to collaborate from outcomes.} We introduce DAG-GRPO, which trains a whole branching episode in one pass, and show that a small model learns when to use each operation from outcome rewards alone (\Cref{sec:method-dag-grpo,sec:experiments}).
\end{itemize}

\section{Octopus: Three Operations on Shared Memory}
\label{sec:method}

Octopus runs a task as a tree of branches served by one model (\Cref{fig:overview}c). Each branch has a context of its own, and all of them draw on one pool of memory. A task starts as one root thread. Any thread can hand off work by forking: it names one or more sub-tasks, and each becomes a child branch, a worker. Workers run in parallel and can fork in turn, so the shape of the team is decided during the task and not before it. A worker ends by reporting an answer with its evidence, and a parent resumes once its children have reported. Three operations act on the shared memory (\Cref{tab:operations}). We describe each in turn, and then the runtime that makes them cheap.
\label{sec:method-framework}

\begin{table}[!htb]
\centering
\caption{The three memory operations. None of them writes a new KV page.}
\label{tab:operations}
\setlength{\tabcolsep}{5pt}
\setlength{\aboverulesep}{0pt}
\setlength{\belowrulesep}{0pt}
\renewcommand{\arraystretch}{1.10}
\small
\begin{tabularx}{\textwidth}{@{}ll>{\hsize=1.3\hsize}X>{\hsize=0.7\hsize}X@{}}
\toprule
\rowcolor{hdrbg}\rowstrut\textbf{Decision} & \textbf{Operation} & \textbf{What happens to memory} & \textbf{What it costs} \\
\midrule
\rowstrut Hand off & \texttt{fork} & Each child shares the parent's keys and values by reference and copies its recurrent state. & Same for any parent length. \\
\rowstrut Forget & \texttt{rewind} & The pages after an earlier step are dropped, the recurrent state of that step is restored, and a note is kept. & Nothing; frees context room. \\
\rowstrut Consult & \texttt{read} & Another branch's pages are mounted into the reader's attention layers; its recurrent layers skip them. & None of the reader's window. \\
\bottomrule
\end{tabularx}
\end{table}

\paragraph{Handing off.}
The usual way to hand off work is to start a sub-agent from a written description of the task, and the sub-agent then knows only what the description says. A fork starts the child from the parent's state instead. The child knows everything the parent knew when it forked, and the sub-task is simply its next instruction. This would be costly if the parent's context had to be copied, and it is not. On a hybrid model a branch's memory has two parts, the keys and values of its attention layers and the state of its recurrent layers. A child shares the first by reference and copies the second, which has a fixed size, so a fork costs the same however long the parent is. Each branch has a token budget of its own, and what it inherits does not count against it.

\paragraph{Forgetting.}
A failed attempt can be left in the context, where it takes the room that later steps need, or it can be compacted by a rule that the model does not control. A rewind lets the thread decide for itself: it goes back to one of its own earlier steps, and everything after that step leaves its context. On a hybrid model removing the tokens is not enough, because the recurrent state has already absorbed them, so the runtime restores the state it saved at that step. The thread keeps one thing, a short note it writes about what it tried and why it failed, so that the lesson stays and the attempt does not. A rewind writes nothing else, and the memory that only the dropped steps held is released.

\begin{figure}[!htb]
\centering
\safefig[width=1.0\textwidth]{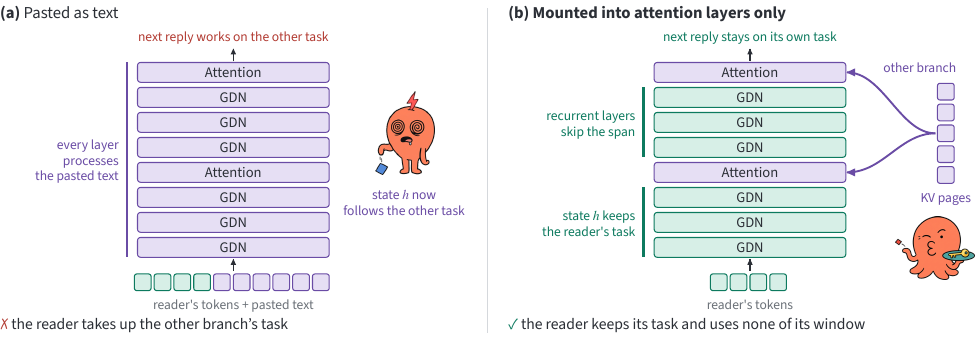}
\caption{Two ways to give a reader another branch's work on a hybrid model of recurrent (GDN) and attention layers; $h$ is the recurrent state.}
\label{fig:yoco}
\end{figure}

\paragraph{Consulting.}
\label{sec:method-cursor-mount}
The usual way to consult another branch is to paste its work into the reader's prompt. Apart from the window this takes, on a hybrid model it derails the reader: more than half of the replies that follow work on the other branch's task and not on the reader's own (\Cref{sec:exp-components}). The cause is in the recurrent layers (\Cref{fig:yoco}a). A recurrent layer folds every token it sees into a fixed-size state, so a long stretch of another branch's work leaves little of the reader's own task behind, whereas an attention layer stores keys and values and reads them only when queried.

A read in Octopus therefore mounts the work into attention layers only (\Cref{fig:yoco}b). The reader's attention layers attend to the other branch's keys and values as if they were part of the reader's own past, while its recurrent layers skip them and carry on from where the reader left off. What attention retrieves still reaches later recurrent layers through the residual stream, so the reader can use what it finds without replaying the other branch's task. One detail needs care. If the other branch has run further than the reader, its keys sit ahead of the reader's query, where causal attention was never trained to look. KVCOMM relocates a copy of the keys into each reader's frame~\citep{ye2025kvcomm}, and Hogwild! rotates the reader's queries instead~\citep{rodionov2025hogwild}. We leave the pages where they are and move the reader: after a read, its next token takes the first position past everything it has mounted (Appendix~\ref{app:runtime-details}).

A read built this way uses none of the reader's window and writes no new memory, so a thread can consult far more than its window could hold. \Cref{sec:exp-components} measures how much a read keeps and how often it still derails the reader, and Appendix~\ref{app:arch-blueprints} carries the same mount to dense and Mamba-2 hybrid models.

\paragraph{Runtime.}
\label{sec:method-kernels-serving}
The three operations are cheap because of how the server keeps memory. Each completed turn is stored as an immutable cursor over reference-counted KV pages on the GPU, extending RadixAttention~\citep{zheng2024sglang}, and a branch's context is a mapping from its logical window to those pages, as a page table is for a process. A fork, a rewind, and a read only change which pages a cursor refers to. None of them writes a page, and a turn sends a cursor and its new messages instead of resending the prompt. The model controls the operations through three tool calls, \texttt{fork}, \texttt{rewind}, and \texttt{read\_thread}, which sit beside the tools of the task itself. Appendix~\ref{app:runtime-details} gives the interface (\Cref{tab:interface}), shows what each operation does to the pool, and describes admission, eviction, and guardrails.

\section{DAG-GRPO: Learning to Hand Off, Forget, and Consult}
\label{sec:method-dag-grpo}

The three operations do not teach a model when to use them: an untrained model that has them does no better than one that does not (Appendix~\ref{app:exp-details}). Their use has to be learned, and standard GRPO does not fit, because it assumes that an episode is one sequence. Each operation breaks that assumption in its own way. A fork gives several branches a common prefix, and splitting the tree into root-to-leaf sequences would compute that prefix once per leaf and count its tokens once per leaf. A rewind removes a segment from the context, yet the actions in it were taken and should receive credit. A read adds an edge between branches that no single sequence contains. DAG-GRPO trains the tree as it is.

\begin{figure}[!htb]
\centering
\safefig[width=1.0\textwidth]{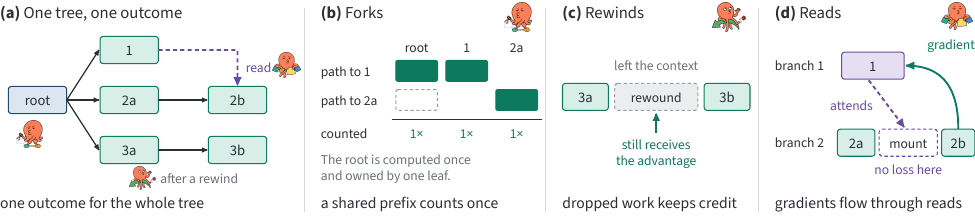}
\caption{How DAG-GRPO trains one episode. Boxes are runs of tokens, and (b) to (d) take their runs from the tree in (a).}
\label{fig:dagrpo}
\end{figure}

The whole tree shares one outcome. For each question we sample $G$ trees and score each by its final answer, $R(\tau_i) = 2y(\tau_i) - 1$ with $y \in \{0, 1\}$. Following REINFORCE Leave-One-Out~\citep{kool2019buy}, the advantage of a tree is its reward minus the mean reward of the other trees,
\begin{equation}
\label{eq:rloo-adv}
\hat{A}(\tau_i) = R(\tau_i) - \frac{1}{G - 1} \sum_{j \neq i} R(\tau_j).
\end{equation}
Every generated token in the tree receives this one advantage, whichever branch produced it (\Cref{fig:dagrpo}a). The three operations each ask one more thing of the computation.

\paragraph{Forks.}
\label{sec:method-kernels}
Handing off work leaves branches with a shared prefix, which has to be computed once and counted once. The episode is cut at every fork, rewind, and read into runs of tokens. Each run is computed once however many branches descend from it, starting from its parent's recurrent state, and the runs are packed into one sequence for a single forward and backward pass. This trains $8.1\times$ more generated tokens per step. Counting needs the same care, because a token on a shared prefix lies on the path to several leaves. Each generated token is therefore assigned to the single leaf that first produced it:
\begin{equation}
\label{eq:single-owner-mask}
m_{\ell,t} = \mathbb{I}\big[\ell = \mathrm{owner}(u_{\ell,t}) \wedge x_{\ell,t} \in \mathcal{V}_{\text{gen}}(\tau_i)\big],
\end{equation}
where $x_{\ell,t}$ is the $t$-th token on the path to leaf $\ell$, $u_{\ell,t}$ is the run that holds it, and $\mathcal{V}_{\text{gen}}$ is the set of tokens the model generated. The masked sum over all leaves then equals the gradient of the log-probability of the whole tree (\Cref{fig:dagrpo}b),
\begin{equation}
\label{eq:tree-score-identity}
\sum_{\ell \in \tau_i} \sum_{t} m_{\ell,t} \nabla_\theta \log \pi_\theta(x_{\ell,t} \mid \mathcal{G}_{\ell,<t}) = \nabla_\theta \log \mathbb{P}_\theta(\tau_i),
\end{equation}
so a tree is trained as one sample and not as one sample per leaf.

\paragraph{Rewinds.}
Forgetting must not erase credit (\Cref{fig:dagrpo}c). A rewound segment is gone from the thread's context but not from the episode. It stays in the training tree as a run of its own, and its tokens receive the tree's advantage like any others. The decision to abandon a line of work, and the note written about it, are therefore credited with how the whole task turned out.

\paragraph{Reads.}
Consulting adds an edge between two branches, and the edge stays in the pass (\Cref{fig:dagrpo}d). A mounted span is not an action of the reader, so it carries zero mask. It is still present in the pass: the reader's attention layers see the keys and values they saw in the rollout, and its recurrent layers skip them as they did then, which lowers the rollout--training KL by $32\times$ against a pass that drops the mounts. Those keys and values come from trainable projections and are not detached, so the reader's loss flows back into how the other branch's memory was written. Reading and being read are trained together, from the same outcome.

We optimize the clipped form of this objective with the usual policy ratio. Appendix~\ref{app:runtime-details} gives the objective and the packing construction step by step (\Cref{fig:dag-pack,fig:rloo}).

\section{Experiments}
\label{sec:experiments}
\label{sec:exp-setup}

We train on OpenResearcher questions~\citep{li2026openresearcher} and test on BrowseComp-Plus~\citep{chen2025browsecompplus}, a benchmark of $830$ deep-search questions over a fixed corpus, searched with BM25. The policy is Qwen3.5-4B with a LoRA adapter, trained with DAG-GRPO for one pass over the training set, and the only reward is whether the final answer is correct. Every thread may add at most $32\text{K}$ tokens to the context it starts from, and a task may create at most ten branches. Appendix~\ref{app:exp-details} gives the full configuration.

\subsection{The Model Learns to Manage Its Memory}
\label{sec:exp-rl}

From outcome rewards alone, a $4$B model learns to use the three operations and reaches the accuracy of OpenAI o3 (\Cref{tab:main-search}, \Cref{fig:pareto}). The operations do nothing by themselves: given to the untrained model, they leave its accuracy where ReAct has it. Nor is the gain in the trained weights alone, since the same weights without \texttt{fork} fall to $30.2\%$ (\Cref{tab:components}). With training and the operations together, Octopus answers $48.3\%$ of the questions. The closest comparison is Single-Agent GRPO, one thread trained with the same rewards and the same recipe in a $143\text{K}$ window. Octopus is ahead of it by $3.4$ points while each of its threads adds at most $32\text{K}$ tokens of its own.

\begin{table}[!htb]
\centering
\caption{Qwen3.5-4B on the $830$ questions of BrowseComp-Plus, with the same tools and grader in every row. Budget is the number of tokens a thread may add to its context; a branch also inherits its parent's context. The last three columns are the shares of a question's evidence documents that search returned and that the agent opened, and the share of its gold documents that the agent opened. Three independent evaluations of Octopus give $48.2 \pm 1.3$. Its lead over Single-Agent GRPO is $3.44$ points against the mean of that agent's eight samples (paired by question, $95\%$ interval $[+0.95, +5.90]$). For reference, OpenAI o3 and GPT-5 score $49.3\%$ and $55.9\%$ with BM25 search~\citep{chen2025browsecompplus}.}
\label{tab:main-search}
\setlength{\tabcolsep}{5pt}
\setlength{\aboverulesep}{0pt}
\setlength{\belowrulesep}{0pt}
\renewcommand{\arraystretch}{1.10}
\small
\begin{tabularx}{\textwidth}{@{}Xllrrrr@{}}
\toprule
\rowcolor{hdrbg} & & & & \multicolumn{3}{c}{\rowstrut\textbf{Documents (\%)}} \\
\rowcolor{hdrbg}\rowstrut\textbf{Agent} & \textbf{Trained} & \textbf{Budget} & \textbf{Accuracy (\%)} & \textbf{found} & \textbf{opened} & \textbf{gold opened} \\
\midrule
\rowstrut ReAct & no & $32\text{K}$ & $26.0$ & $28.6$ & $7.5$ & $12.2$ \\
\rowstrut The three operations, not trained & no & $32\text{K}$ per thread & $26.3$ & $30.5$ & $8.6$ & $12.5$ \\
\rowstrut Single-Agent GRPO & yes & $143\text{K}$ & $44.9$ & $43.9$ & $10.2$ & $17.3$ \\
\rowcolor{oursrowdeep}\rowstrut\octodense{octo_21_title_boss}\textbf{Octopus} & yes & $32\text{K}$ per thread & $\mathbf{48.3}$ & $\mathbf{57.4}$ & $\mathbf{35.7}$ & $\mathbf{43.3}$ \\
\bottomrule
\end{tabularx}
\end{table}

\label{sec:exp-context}
The two reach their answers differently. Single-Agent GRPO finds much of a question's evidence by search and opens little of it. Octopus finds more and reads it: it opens $43.3\%$ of the gold documents, against $17.3\%$, and $97.0\%$ of its correct answers opened a gold document, against $59.3\%$. A long window lets an agent search widely, and separate windows let it read what it finds. The tree also does more work than one thread, and Appendix~\ref{app:compute-curve} compares the two at matched inference cost.

\begin{figure}[!htb]
\centering
\safefig[width=1.0\textwidth]{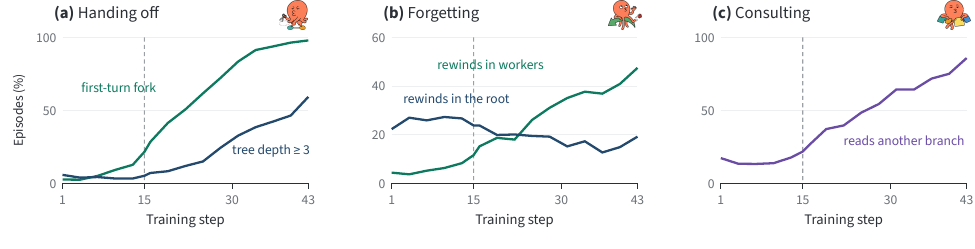}
\caption{When the policy learns each operation: the share of training episodes that use it, over the $24{,}041$ episodes of the training run. The dashed line marks step $15$.}
\label{fig:learned}
\end{figure}

\label{sec:exp-learned}
No reward term names an operation, and the policy learns all three (\Cref{fig:learned}). They take hold together around training step $15$. By the end of training nearly every episode hands off work on its first turn, rewinds have moved from the root to the workers, and most episodes consult another branch. The policy also finds a use for forks beyond dividing work: in $26$ of $48$ validation rollouts it forks one task at a time, up to ten levels deep, and uses each fork as a fresh window. A second training run of our own, limited to one level of branching, rises from $19.8\%$ to $42.3\%$ on the same benchmark (Appendix~\ref{app:exp-details}).

\subsection{Ablations: What Each Operation Contributes}
\label{sec:exp-components}
\label{sec:exp-scaling}

\Cref{tab:components} removes or limits one operation of the trained policy at test time. Handing off is worth the most. Accuracy falls as fewer branches are allowed, slowly down to six and then faster, to $37.7\%$ with two, and without \texttt{fork} the policy loses $18$ points.

\begin{table}[!htb]
\centering
\caption{What each operation contributes: the trained $4$B policy with one operation removed or limited at test time, on BrowseComp-Plus. The last column is the accuracy lost against the full policy, in points, paired by question, with its $95\%$ bootstrap interval.}
\label{tab:components}
\setlength{\tabcolsep}{5pt}
\setlength{\aboverulesep}{0pt}
\setlength{\belowrulesep}{0pt}
\renewcommand{\arraystretch}{1.10}
\small
\begin{tabularx}{\textwidth}{@{}lXrr@{}}
\toprule
\rowcolor{hdrbg}\rowstrut\textbf{Operation} & \textbf{Policy at test time} & \textbf{Accuracy (\%)} & \textbf{Accuracy lost} \\
\midrule
\rowcolor{oursrowdeep}\rowstrut & \octodense{octo_21_title_boss}\textbf{Octopus} (all three, up to ten branches) & $\mathbf{48.3}$ & --- \\
\rowstrut Hand off & At most eight branches & $47.6$ & $+0.72$ [$-2.05$, $+3.62$] \\
\rowstrut & At most six branches & $46.5$ & $+1.81$ [$-1.09$, $+4.72$] \\
\rowstrut & At most four branches & $42.7$ & $+5.55$ [$+2.65$, $+8.44$] \\
\rowstrut & At most two branches & $37.7$ & $+10.62$ [$+7.72$, $+13.63$] \\
\rowstrut & \texttt{fork} removed & $30.2$ & $+18.09$ [$+15.08$, $+21.11$] \\
\rowstrut Forget & \texttt{rewind} removed & $42.3$ & $+5.93$ [$+3.02$, $+8.83$] \\
\rowstrut Consult & \texttt{read\_thread} removed & $46.3$ & $+2.05$ [$-0.72$, $+5.07$] \\
\bottomrule
\end{tabularx}
\end{table}

Forgetting is used in nine of every ten test episodes, and without \texttt{rewind} the policy loses $5.9$ points.

\label{sec:exp-mechanism}
For consulting we test the read itself (\Cref{tab:kv-probe}). At the points in real rollouts where one branch read another, we regenerate the reader's next reply under each way of delivering the other branch's steps, and a judge labels whether that reply works on the reader's own task or on the other branch's. With nothing delivered, $0.5\%$ of replies work on the other branch's task. Pasting the steps as text raises this to $57.9\%$. Splicing the other branch's cached state into every layer pastes no text and does the same, so the failure follows the recurrent layers and not the text. Mounting into attention layers alone brings it back to $2.7\%$ while keeping $92.2\%$ of the answer information of the full text, and it adds no KV memory. Text could not have carried these reads in any case: a read has a median of $18.3\text{K}$ tokens, and $44\%$ of the branches that read would need more than a whole window to hold what they read.\footnote{The trained policy on $48$ OpenResearcher validation questions, $162$ reading branches.} Removing reads from the trained policy changes accuracy by $2.1$ points, which is within noise on this benchmark.

\begin{table}[!htb]
\centering
\caption{How a branch should read another: six ways to deliver another branch's steps, measured on Qwen3.5-4B at $366$ reads taken from real rollouts. Off task is the share of the reader's next replies that work on the other branch's task. Appendix~\ref{app:exp-details} adds recall and policy drift.}
\label{tab:kv-probe}
\setlength{\tabcolsep}{5pt}
\setlength{\aboverulesep}{0pt}
\setlength{\belowrulesep}{0pt}
\renewcommand{\arraystretch}{1.10}
\small
\begin{tabularx}{\textwidth}{@{}Xrrr@{}}
\toprule
\rowcolor{hdrbg}\rowstrut\textbf{How the other branch is read} & \textbf{Extra KV memory} & \textbf{Information kept} & \textbf{Off task} \\
\midrule
\rowstrut Not read & $0$ & $0.0\%$ & $0.5\%$ \\
\rowstrut Pasted as text~\citep{wu2023autogen} & $+100\%$ & $100\%$ & $57.9\%$ \\
\rowstrut KV spliced into all layers~\citep{yao2025cacheblend} & $+100\%$ & $98.1\%$ & $55.2\%$ \\
\rowstrut KV copied and rotated~\citep{ye2025kvcomm} & $+100\%$ & $91.7\%$ & $7.1\%$ \\
\rowstrut Mounted into attention layers, positions unchanged & $0$ & $86.1\%$ & $2.5\%$ \\
\rowcolor{oursrowdeep}\rowstrut\octodense{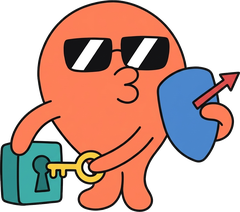}\textbf{Mounted into attention layers, reader moved (Octopus)} & $\mathbf{0}$ & $\mathbf{92.2\%}$ & $\mathbf{2.7\%}$ \\
\bottomrule
\end{tabularx}
\end{table}

\subsection{A Larger Model without Training}
\label{sec:exp-zeroshot}

A small model has to be trained to use the operations. A larger one can use them from a description. We serve Qwen3.6-27B, a hybrid of the same family with $48$ recurrent and $16$ attention layers, through the same runtime and do not train it. A short prompt describes the three operations and when each is useful. We compare it with three single agents on the same model, each with a strategy prompt of its own: one thread in a $32\text{K}$ window, one thread in a long window, and seven independent $32\text{K}$ runs from which the model picks one answer. The last two spend about as much inference as Octopus (Appendix~\ref{app:exp-details}).

\begin{table}[!htb]
\centering
\caption{Qwen3.6-27B on BrowseComp-Plus with no training. Each agent has a strategy prompt written for it. The last column is the lead of Octopus in points, paired by question, with its $95\%$ bootstrap interval.}
\label{tab:zeroshot}
\setlength{\tabcolsep}{5pt}
\setlength{\aboverulesep}{0pt}
\setlength{\belowrulesep}{0pt}
\renewcommand{\arraystretch}{1.10}
\small
\begin{tabularx}{\textwidth}{@{}Xrr@{}}
\toprule
\rowcolor{hdrbg}\rowstrut\textbf{Agent} & \textbf{Accuracy (\%)} & \textbf{Octopus leads by} \\
\midrule
\rowstrut One thread, $32\text{K}$ window & $36.8$ & $+28.19$ [$+24.94$, $+31.57$] \\
\rowstrut One thread, long window & $56.1$ & $+8.80$ [$+5.66$, $+11.81$] \\
\rowstrut Seven $32\text{K}$ runs, the model picks one answer & $52.7$ & $+12.29$ [$+9.40$, $+15.18$] \\
\rowcolor{oursrowdeep}\rowstrut\octodense{octo_21_title_boss}\textbf{Octopus} & $\mathbf{64.9}$ & --- \\
\bottomrule
\end{tabularx}
\end{table}

Octopus reaches $64.9\%$ and leads all three (\Cref{tab:zeroshot}), by $8.8$ points over the long window and by $12.3$ over the seven runs. A larger model can therefore use the runtime with no training.

\section{Related Work}
\label{sec:related}

Earlier work addresses each of the three decisions, and \Cref{tab:related-comparison} places representative systems along them.

\begin{table}[!htb]
\centering
\caption{Representative systems along the three decisions. \emph{Hands off}: the model decides at run time to start new branches or agents. \emph{Forgets}: the model decides to drop part of its own context. \emph{Consults}: a branch or agent can read the work of a peer, beyond a result returned to the level above. \emph{Shared as}: what passes between contexts. \emph{Learned}: these decisions are trained from the reward of the task (\cmarkyes~yes, \cmarkpart~partly, \cmarkno~no).}
\label{tab:related-comparison}
\setlength{\tabcolsep}{5pt}
\setlength{\aboverulesep}{0pt}
\setlength{\belowrulesep}{0pt}
\renewcommand{\arraystretch}{1.10}
\small
\begin{tabularx}{\textwidth}{@{}Xccclc@{}}
\toprule
\rowcolor{hdrbg}\rowstrut\textbf{System} & \textbf{Hands off} & \textbf{Forgets} & \textbf{Consults} & \textbf{Shared as} & \textbf{Learned} \\
\midrule
\rowstrut AutoGen~\citep{wu2023autogen} & \cmarkno & \cmarkno & \cmarkyes & text & \cmarkno \\
\rowstrut WideSeek-R1~\citep{xu2026wideseekr1} & \cmarkyes & \cmarkno & \cmarkno & text & \cmarkyes \\
\rowstrut APR~\citep{pan2025apr} & \cmarkyes & \cmarkno & \cmarkno & text & \cmarkyes \\
\rowstrut ThreadWeaver~\citep{lian2025threadweaver} & \cmarkyes & \cmarkno & \cmarkno & text & \cmarkyes \\
\rowstrut Context-Folding~\citep{sun2025contextfolding} & \cmarkyes & \cmarkyes & \cmarkno & text & \cmarkyes \\
\rowstrut CLM~\citep{shao2026clm} & \cmarkpart & \cmarkyes & \cmarkyes & text & \cmarkyes \\
\rowstrut Hogwild!~\citep{rodionov2025hogwild} & \cmarkno & \cmarkno & \cmarkyes & KV & \cmarkno \\
\rowstrut KVCOMM~\citep{ye2025kvcomm} & \cmarkno & \cmarkno & \cmarkyes & KV & \cmarkno \\
\rowcolor{oursrowdeep}\rowstrut\octodense{octo_21_title_boss}\textbf{Octopus} & \cmarkyes & \cmarkyes & \cmarkyes & KV & \cmarkyes \\
\bottomrule
\end{tabularx}
\end{table}

\paragraph{Managing one context.}
A single thread stays usable if part of its history is dropped or moved out of the window, and systems differ in who decides. ReSum summarizes when the window is nearly full, by a rule of the system~\citep{wu2025resum}. MEM1 and MemAgent rewrite a fixed memory on every turn or segment~\citep{zhou2026mem1,yu2026memagent}. MemGPT, AgentFold, ACM, Context-Folding, and CLM let the model choose when to page out, fold, compress, or edit its context~\citep{packer2023memgpt,ye2026agentfold,li2026acm,sun2025contextfolding,shao2026clm}. A rewind in Octopus returns a thread to an earlier point and leaves a note. It restores that point exactly on a hybrid model, recurrent state included, and it is learned together with handing off and consulting.

\paragraph{Handing work to other agents or branches.}
Multi-agent frameworks such as AutoGen, CAMEL, and MetaGPT have agents exchange messages along patterns that their designer writes~\citep{wu2023autogen,li2023camel,hong2023metagpt}. Later systems train the hand-off. WideSeek-R1, WideSeek, and M-GRPO train a lead agent that decides during the task what to delegate~\citep{xu2026wideseekr1,huang2026wideseek,hong2025mgrpo}, and APR, Parallel-R1, ThreadWeaver, AsyncThink, Multiverse, and NPR train one model to open parallel threads~\citep{pan2025apr,zheng2025parallelr1,lian2025threadweaver,chi2025asyncthink,yang2025multiverse,wu2026npr}. In most of these a thread or sub-agent works without seeing its peers, and its result returns to the level above as text. In Octopus any branch can read any other while both run, and branches fork to any depth.

\paragraph{Sharing memory instead of text.}
Serving systems already share KV memory across requests. PagedAttention and RadixAttention share prefixes~\citep{kwon2023efficient,zheng2024sglang}, and Prompt Cache, CacheBlend, and EPIC reuse cached blocks at new positions~\citep{gim2024promptcache,yao2025cacheblend,hu2025epic}. Between agents, Hogwild! lets workers attend to each other's tokens in one shared cache~\citep{rodionov2025hogwild}, KVCOMM reuses the KV of content that agents have in common~\citep{ye2025kvcomm}, KVComm passes selected layers of a sender's KV to a receiver~\citep{shi2025kvcomm}, and Cache-to-Cache and DiffMAS train models to consume another model's KV~\citep{fu2026c2c,yu2026diffmas}. Other work passes activations or latent states between agents~\citep{ramesh2025communicating,zou2026latentmas}. These methods act on attention or on hidden states, with a pattern of sharing that the designer fixes. Octopus targets hybrid models, whose recurrent layers hold a fixed-size state that shapes how the model continues and cannot be reused as a partial prefix~\citep{yang2024gated,afendulev2026attention,jelassi2024repeat,pan2025marconi}. It mounts another branch into the attention layers alone, lets the model choose when to read, and trains the read with the rest of the policy.

\paragraph{Training over trees.}
Tree Training and Prefix Grouper compute the shared prefixes of a set of trajectories once~\citep{wang2025treetraining,liu2025prefixgrouper}. The tree that DAG-GRPO trains also holds recurrent state, rewound segments, and mounted spans. Tree-GRPO, TreeRL, and ARPO also train on trees, whose branches are alternative continuations of one question and are compared with each other~\citep{ji2025treegrpo,hou2025treerl,dong2026arpo}; in Octopus the tree is what the agent builds within one episode, and its branches work toward one answer. CCPO and C3 estimate the share of a team's reward that belongs to each role or each message with counterfactual rollouts~\citep{li2026ccpo,chen2026c3}, whereas DAG-GRPO gives every token of a tree the same outcome advantage.

\section{Conclusion}
\label{sec:discussion}

A long task asks a model to manage its working memory, and we asked whether the model can do this itself. Octopus gives it three operations on memory that its branches share: handing off work, forgetting a dead end, and consulting another branch without losing its own task. Trained on nothing but whether the final answer was right, a $4$B model learns when to use each of them and reaches the accuracy of OpenAI o3. The decisions that agent systems now make for a model are decisions the model can learn to make.

\paragraph{Limitations and future work.}
We evaluate Octopus on one kind of task, deep search, and on one family of hybrid models, and the attention-only read is designed for models with recurrent layers. On BrowseComp-Plus, removing reads does not change accuracy measurably. We expect consulting to matter most where branches depend on one another's work, as in software engineering, where they edit one codebase, and such tasks, together with broader search benchmarks such as WideSearch~\citep{wong2025widesearch}, are the natural next tests. A tree also does more work than a single thread, as other ways of scaling test-time compute do (Appendix~\ref{app:compute-curve} reports the costs), and our reward has no term for cost. Adding one is a natural next step.

\bibliographystyle{unsrtnat}
\bibliography{references_v2}

\appendix
\setlength{\parskip}{3.5pt plus 0.6pt minus 0.3pt}

\section{Runtime and Training Details}
\label{app:runtime-details}

\begin{table}[!htb]
\centering
\caption{The tool calls through which the model controls the three memory operations. A branch ends with the closing call of its task, \texttt{done(answer, evidence)} in search, and a parent resumes when all of its children have ended.}
\label{tab:interface}
\setlength{\tabcolsep}{5pt}
\setlength{\aboverulesep}{0pt}
\setlength{\belowrulesep}{0pt}
\renewcommand{\arraystretch}{1.10}
\small
\begin{tabularx}{\textwidth}{@{}llX@{}}
\toprule
\rowcolor{hdrbg}\rowstrut\textbf{Operation} & \textbf{Tool call} & \textbf{Effect} \\
\midrule
\rowstrut Hand off & \texttt{fork(tasks)} & Starts one branch per task. Each shares the caller's KV pages and copies its recurrent state. \\
\rowstrut Forget & \texttt{rewind(to\_step, note)} & Returns the caller to the end of an earlier step. Later steps leave its context and the note takes their place. \\
\rowstrut Consult & \texttt{read\_thread(thread, from\_step, to\_step)} & Mounts those steps of another thread into the caller's attention layers from its next reply on. \\
\bottomrule
\end{tabularx}
\end{table}

\begin{figure}[!htb]
\centering
\safefig[width=1.0\textwidth]{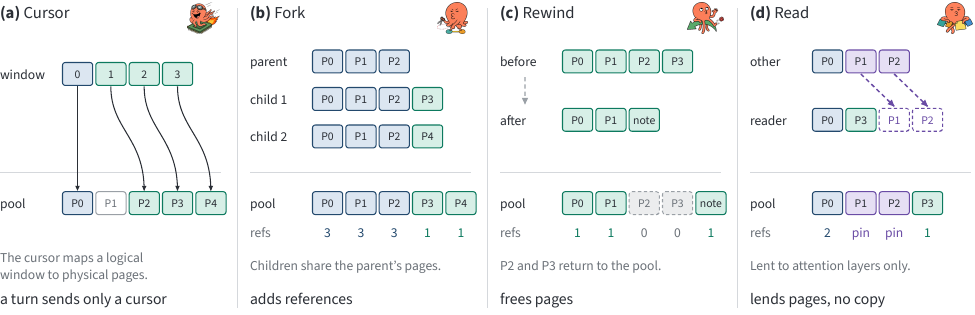}
\caption{One pool of reference-counted KV pages. Each row lists the pages a cursor refers to, and refs is the reference count of each page.}
\label{fig:paged}
\end{figure}

\paragraph{Pages and cursors.}
\Cref{fig:paged} shows the pool before and after each operation. A fork adds references to the parent's pages, a rewind returns the pages that no other cursor holds, and a read pins another branch's pages and lends them to the reader's attention layers. None of the three writes a page.

\begin{figure}[t]
\centering
\safefig[width=1.0\textwidth]{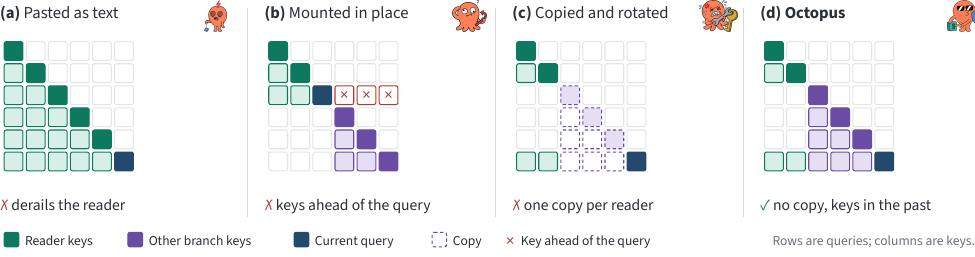}
\caption{Where another branch's keys sit relative to the reader's query under four ways of reading it.}
\label{fig:rope-streaming}
\end{figure}

\paragraph{Position rule.}
After a read, the reader's next token takes the first position past everything it has mounted,
\begin{equation}
\label{eq:k2-pos}
s(c') = \max\!\Big(\mathrm{pos}_{\text{next}}(c),\; \max_{m \in \mathrm{Mounts}(c')} \big(\mathrm{pos}_{\text{end}}(m) + 1\big)\Big),
\end{equation}
where $c$ and $c'$ are the reader's cursor before and after the read, $s(c')$ is the position of its next token, $\mathrm{pos}_{\text{next}}(c)$ is where that token would otherwise have gone, and $m$ ranges over the spans the reader has mounted. All mounted keys then lie in the reader's past, and no page is copied (\Cref{fig:rope-streaming}).

\paragraph{Packed pass.}
An episode compiles into one forward--backward pass in three steps.
\begin{enumerate}
  \item \textbf{Split.} The episode is cut at every \texttt{fork}, \texttt{rewind}, and read boundary into maximal token runs; each run becomes one node with a tree parent, its mounted spans, and a start position past everything it mounts, evaluated once however many branches descend from it.
  \item \textbf{Pack.} The nodes are concatenated topologically into one unpadded tensor, and all MLP and projection layers run over it in one pass.
  \item \textbf{Route.} Recurrent layers start each node from its parent's final state and skip mounted spans. Each attention layer that mounts gathers its keys from the mounted spans, the ancestors, and the node itself:
\end{enumerate}
\begin{equation}
\label{eq:dag-attn}
\mathbf{K}_u^{(l)} = \Big[\, \bigoplus_{m \in \mathcal{M}(u)} \mathbf{K}_m^{(l)} \;\Big\|\; \mathbf{K}_{p(u)}^{(l)} \;\Big\|\; \mathbf{K}_{\text{self}(u)}^{(l)} \,\Big].
\end{equation}

\begin{figure}[!htb]
\centering
\safefig[width=1.0\textwidth]{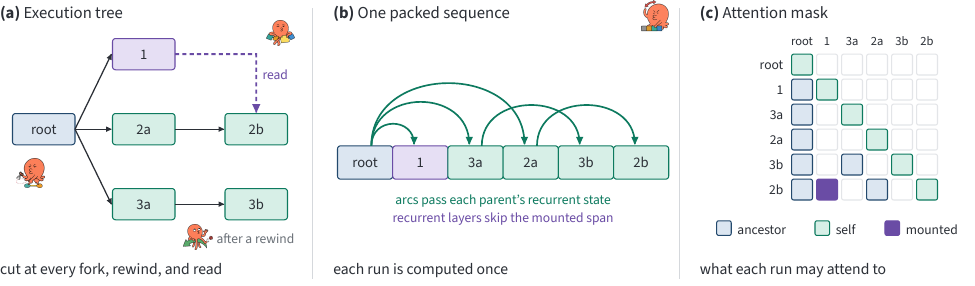}
\caption{Training one episode in one pass. (a) The tree is cut into runs at every fork, rewind, and read: branch 2 reads branch 1 between runs 2a and 2b, and branch 3 rewinds between 3a and 3b. (b) Each run is packed once into one sequence; arcs pass each parent's recurrent state to its children, and recurrent layers skip mounted spans. (c) An attention layer sees a run's ancestors, the run itself, and its mounted spans.}
\label{fig:dag-pack}
\end{figure}
\begin{figure}[!htb]
\centering
\safefig[width=1.0\textwidth]{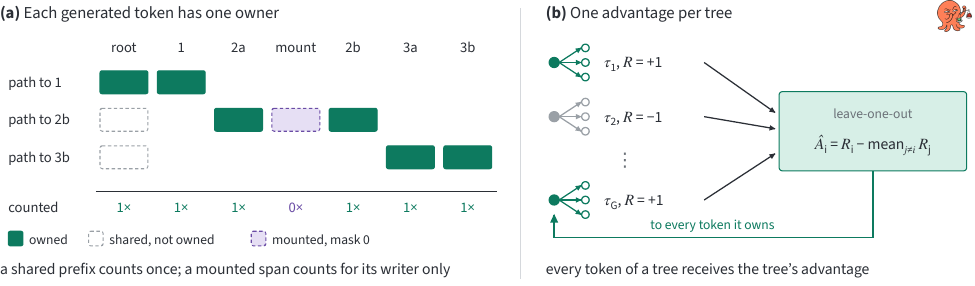}
\caption{Credit assignment in DAG-GRPO, for the tree of \Cref{fig:dag-pack}. (a) Each generated token is owned by the first leaf whose path contains it, so a shared prefix counts once and a mounted span is not counted for its reader. (b) One leave-one-out advantage per tree goes to every token the tree owns.}
\label{fig:rloo}
\end{figure}
\paragraph{Objective.}
DAG-GRPO maximizes the clipped surrogate
\begin{equation}
\label{eq:dag-grpo}
\mathcal{J}(\theta) = \mathbb{E}\! \left[ \frac{1}{G} \sum_{i=1}^G \frac{1}{Z_i} \sum_{\ell \in \tau_i} \sum_{t=1}^{|\ell|} m_{\ell,t} \, \min\!\Big( r_{\ell,t}(\theta) \hat{A}(\tau_i),\; \mathrm{clip}\big(r_{\ell,t}(\theta), 1-\epsilon, 1+\epsilon\big) \hat{A}(\tau_i) \Big) \right],
\end{equation}
where $Z_i = \sum_{\ell,t} m_{\ell,t}$ and $r_{\ell,t}(\theta)$ is the policy ratio. The factor $1/Z_i$ keeps gradient magnitudes comparable across trees that span $1\text{K}$ to $199\text{K}$ tokens.

\paragraph{Chat templates.}
Open-weight chat templates often break append-only rendering: Qwen2.5/3/3.5 strip earlier assistant reasoning on a new user turn, DeepSeek-V3/R1 format the first and later tool calls differently, and gpt-oss drops analysis channels when a conversation is rendered again. At server startup (about $0.1\,\text{s}$) the runtime renders synthetic pairs of a parent reply and a following turn, locates the token that ends an assistant turn, and classifies the template into one of two incremental modes. Tokenization is then bit-exact across $18{,}344$ test turns. Runtime notes are delivered as tagged tool responses, so earlier reasoning blocks stay intact.

\paragraph{Tool schema, answer reserve, and eviction.}
The read tool is left out of the static system prompt. Its schema and the list of other branches are appended only after another branch has completed a step, which keeps the shared prefix intact, and requested step ranges are clamped to the steps the source has. Within each thread's budget, $2{,}048$ tokens are reserved for the closing call. Mounted spans are held through reference-counted page pointers; if memory pressure or an online LoRA update evicts one, the server prefills it again from the cursor's offsets before decoding.

\paragraph{Admission.}
New episodes are admitted only while the KV pool is below $75\%$ full, so that a burst of forks in the middle of an episode cannot exhaust GPU memory. Without the gate, the 90th-percentile pool utilization reaches $98\%$ to $99\%$ and $14.7\%$ to $46.1\%$ of episodes time out; with it, the 90th percentile is $74.5\%$, with no queued requests and no timeouts across $1{,}304$ episodes. Each $141\text{GB}$ GPU holds a pool of $1.94\text{M}$ tokens, up to $524$ saved recurrent states, and $154$ threads decoding at once. A branch may take up to $50$ turns; at $20$, branches that had rewound ran out of turns before their closing call.

\paragraph{Termination prompt.}
A reply without a tool call is answered once with a prompt to finish, and that prompt is masked from the loss. Training does not hold without it. Across $33{,}075$ pooled rollouts, when the answer is extracted from a root's plain-text reply, such replies rise from $6\%$ to $75\%$ of endings by step $12$. When only a \texttt{done} call is graded and no prompt is given, the $7.5\%$ of forking roots that answer in plain text after their branches return receive zero reward, forking is penalized, and the share of roots that fork falls from $23.9\%$--$28.2\%$ to $0\%$ by step $15$. With the prompt, $97.4\%$ of prompted roots and $93.1\%$ of prompted branches close with a valid \texttt{done}, and $98.8\%$ to $99.2\%$ of roots fork throughout the pass.

\section{Experimental Details}
\label{app:exp-details}

\begin{table}[!htb]
\centering
\caption{Configuration of the experiments in \Cref{sec:experiments}.}
\label{tab:config}
\setlength{\tabcolsep}{5pt}
\setlength{\aboverulesep}{0pt}
\setlength{\belowrulesep}{0pt}
\renewcommand{\arraystretch}{1.10}
\small
\begin{tabularx}{\textwidth}{@{}llX@{}}
\toprule
\rowstrut\textbf{Group} & \textbf{Setting} & \textbf{Value} \\
\midrule
\rowstrut Data & Training & OpenResearcher, $3{,}040$ questions ($500$ more held out for validation), corpus of $14.9\text{M}$ documents \\
\rowstrut & Test & BrowseComp-Plus, $830$ questions, corpus of $100{,}195$ documents \\
\rowstrut & Search & BM25, ten hits of $512$ tokens per query; \texttt{open} returns pages of $8\text{K}$ tokens \\
\midrule
\rowstrut Models & Trained & Qwen3.5-4B ($24$ recurrent and $8$ attention layers), LoRA of rank $64$ \\
\rowstrut & Not trained & Qwen3.6-27B ($48$ recurrent and $16$ attention layers), thinking disabled \\
\midrule
\rowstrut Training & Algorithm & DAG-GRPO, binary outcome reward, groups with equal rewards dropped \\
\rowstrut & Schedule & $45$ steps (one pass over the training set), $16$ questions per step, $G=8$ episodes per question \\
\midrule
\rowstrut Episode & Thread budget & $32{,}768$ tokens added per thread, of which $2{,}048$ are reserved for the closing call \\
\rowstrut & Tree & Depth unlimited, at most ten branches per task, at most two rewinds per thread \\
\rowstrut & Sampling & Temperature $0.7$, top-$p$ $0.95$ \\
\midrule
\rowstrut Baselines & ReAct & Untrained model, one thread, $32\text{K}$ budget \\
\rowstrut & Single-Agent GRPO & Same reward and recipe, one thread, $143\text{K}$ budget, at most $78$ tool calls, $40$ steps; mean of eight samples \\
\midrule
\rowstrut Evaluation & Grader & Qwen3.5-27B at temperature $0$, following \citet{shao2026clm} \\
\rowstrut & Statistics & Paired by question, $10{,}000$ bootstrap resamples, $95\%$ intervals; three independent evaluations of Octopus \\
\midrule
\rowstrut System & Rollouts & SGLang with the Octopus runtime \\
\rowstrut & Training & AReaL \\
\bottomrule
\end{tabularx}
\end{table}

\paragraph{Data and measurement.}
The training and test corpora are separate: no training question appears in BrowseComp-Plus, and $2$ of $8{,}306$ gold URLs overlap, with accuracy unchanged when those are excluded. Unless a row states another step, 4B results use the checkpoint after one pass. Cost is reported as PF/Q, the inference PFLOPs per question with prefix reuse, summed over requests~\citep{shao2026clm}.\footnote{PF/Q does not yet include attention over mounted keys, except where a table adds it.}

\paragraph{Reading another branch.}
\Cref{tab:kv-full} extends \cref{tab:kv-probe} to every way of reading we tested. Pasting text and splicing KV into all layers both run the reader's recurrent layers over the other branch's tokens. The reader's negative log-likelihood on its own continuation rises by $0.196$ and $0.195$, and $57.9\%$ and $55.2\%$ of its next replies leave its task. Keeping the recurrent layers off the path brings this to $2.5\%$. Without the position rule of \cref{eq:k2-pos}, however, exact-match recall on branches that ran past the reader falls to $0.449$, because their keys land in the reader's future. With the rule, recall on those branches is $0.613$, $92.2\%$ of the answer information in the full text is retained, and no page is added. Over $115$ reads in live rollouts, a blind judge finds no more off-task replies than under a control that blocks the read ($-1.1\%$).

\begin{table}[!htb]
\centering\small
\caption{Reading another branch on Qwen3.5-4B over $366$ live mounts and $947$ exact-match probes. Off-task is the share of follow-up turns that abandon the reader's task; recall is exact match over all branches read and over those that ran past the reader; $\Delta$NLL is policy drift.}
\label{tab:kv-full}
\setlength{\tabcolsep}{5pt}
\setlength{\aboverulesep}{0pt}
\setlength{\belowrulesep}{0pt}
\renewcommand{\arraystretch}{1.10}
\small
\begin{tabularx}{\textwidth}{@{}Xrrrrr@{}}
\toprule
\rowcolor{hdrbg}\rowstrut\textbf{How the other branch is read} & \textbf{Extra KV} & \textbf{Info retained} & \textbf{Recall (all / longer)} & \textbf{$\Delta$NLL} & \textbf{Off-task} \\
\midrule
\rowstrut No read & $0$ & $0.0\%$ & $0.031$ / $0.032$ & $0.000$ & $0.5\%$ \\
\rowstrut Paste as text~\citep{wu2023autogen} & $+100\%$ & $100\%$ & $0.760$ / $0.786$ & $+0.196$ & $57.9\%$ \\
\rowstrut Splice KV into all layers~\citep{yao2025cacheblend} & $+100\%$ & $98.1\%$ & $0.715$ / $0.731$ & $+0.195$ & $55.2\%$ \\
\rowstrut Rotated KV copy~\citep{ye2025kvcomm} & $+100\%$ & $91.7\%$ & $0.601$ / $0.611$ & $+0.015$ & $7.1\%$ \\
\rowstrut Attention layers only, no offset & $0$ & $86.1\%$ & $0.521$ / $0.449$ & $+0.003$ & $2.5\%$ \\
\rowcolor{oursrowdeep}\rowstrut\octodense{octo_k2}\textbf{Attention layers only, with offset (Octopus)} & $\mathbf{0}$ & $\mathbf{92.2\%}$ & $\mathbf{0.612}$ / $\mathbf{0.613}$ & $\mathbf{+0.007}$ & $\mathbf{2.7\%}$ \\
\bottomrule
\end{tabularx}\par\vspace{2.5pt}
\noindent\parbox{\textwidth}{\footnotesize\textbf{Across architectures:} the mount keeps $96.9\%$ argmax agreement with full-context replay on dense Qwen3-4B ($9$ of $36$ layers mounted) and $98.5\%$ token agreement on Mamba-2 hybrid Granite-4.0-H. \textbf{Blind audit:} over $115$ live reads, a judge finds no excess drift against a control that blocks the read ($\Delta$off-task $=-1.1\%$).}
\end{table}

\begin{figure}[!htb]
\centering
\safefig[width=1.0\textwidth]{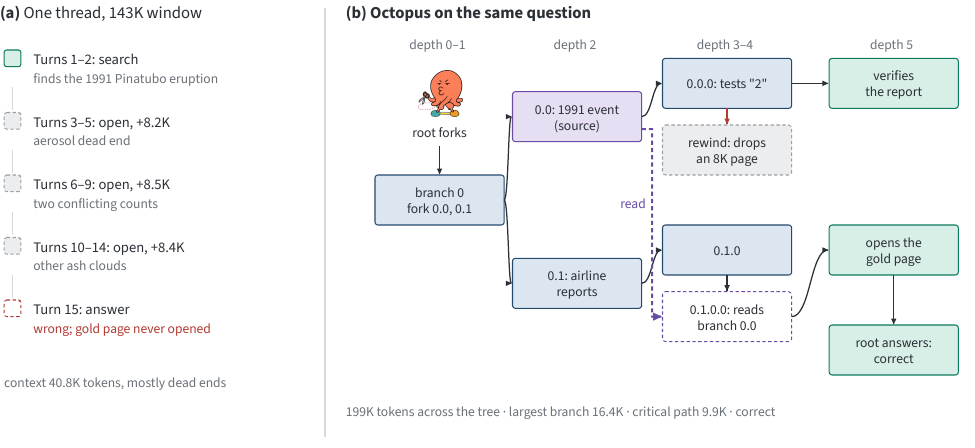}
\caption{One question solved two ways (\texttt{search-3190-99}). (a) One thread with a long window keeps its dead ends in context and answers wrongly. (b) Octopus hands the work to ten branches, one of which rewinds a dead end and another reads a third, and answers correctly.}
\label{fig:rollout}
\end{figure}

\paragraph{An example episode.}
\Cref{fig:rollout} follows one question through a single thread with a long window and through Octopus.

\paragraph{A second training run.}
We trained a second policy ourselves from the base model with the same reward, limited to one level of branching and without \texttt{rewind}, for $40$ steps. On BrowseComp-Plus it rises from $19.8\%$ before training to $42.3\%$, and on OpenResearcher validation from $37.2\%$ to $46.8\%$; the share of episodes that fork rises from $28\%$ to $99\%$. This run used a different grader (Qwen3-32B) and two samples per question, so its numbers are not comparable with \cref{tab:main-search} in absolute terms. An earlier run with the same limit reached $42.6\%$ on BrowseComp-Plus at step $35$.

\paragraph{Learned behaviour.}
Rollouts of the trained policy on $48$ OpenResearcher validation questions show two uses of handing off. In $22$ episodes a fork creates at least two branches. About half of these split the question into different clues, and the branches of one fork retrieve largely different documents (mean pairwise Jaccard $0.28$). In the other $26$ episodes the policy hands off one task at a time, up to ten levels deep, so that each fork gives the work a fresh budget. The policy reads other branches $6.6$ times per episode and rewinds $2.75$ times, almost entirely inside workers. An untrained 27B model with a strategy prompt does so $1.9$ and $0.1$ times on the same questions.

\paragraph{The main comparison in full.}
\Cref{tab:main-full} adds cost and paired differences to \cref{tab:main-search}. Three independent evaluations of Octopus give $48.3\%$, $49.5\%$ and $46.9\%$, and an earlier evaluation of the same checkpoint gave $49.6\%$. Single-Agent GRPO comes from one training run of $40$ steps and was still improving when it stopped: its accuracy at step $20$ is $7.5$ points lower. Octopus is $3.44$ points ahead of the mean of its eight samples and $1.93$ ahead of its first sample. The trained Octopus weights with \texttt{fork} removed score $30.2\%$ in one run, and seven such runs with a majority vote score $41.1\%$ at a higher cost than Octopus. Appendix~\ref{app:compute-curve} compares the agents at equal cost.

\begin{table}[!t]
\centering
\caption{Qwen3.5-4B on BrowseComp-Plus with cost. PF/Q is inference cost per question. The middle rows sample the trained Octopus weights with \texttt{fork} removed: one run, seven runs with a majority vote, seven runs from which the model picks one answer, and seven runs counted correct if any is. The last column is the lead of Octopus in points, paired by question, with its $95\%$ bootstrap interval.}
\label{tab:main-full}
\setlength{\tabcolsep}{6pt}
\renewcommand{\arraystretch}{1.08}
\begin{tabular*}{\textwidth}{@{\extracolsep{\fill}}lccc}
\toprule
\textbf{Agent} & \textbf{Accuracy} & \textbf{PF/Q} & \textbf{Octopus leads by} \\
\midrule
ReAct & $26.0\%$ & $0.33$ & --- \\
\midrule
Single-Agent GRPO, mean of eight samples & $44.9\%$ & $1.10$ & $+3.44$ [$+0.95$, $+5.90$] \\
Single-Agent GRPO, first sample & $46.4\%$ & $1.10$ & $+1.93$ [$-1.21$, $+5.07$] \\
\midrule
\texttt{fork} removed, one run & $30.2\%$ & $0.60$ & $+18.1$ [$+15.1$, $+21.1$] \\
\texttt{fork} removed, seven runs, majority vote & $41.1\%$ & $4.23$ & $+7.2$ [$+4.5$, $+10.1$] \\
\texttt{fork} removed, seven runs, the model picks one & $45.7\%$ & $4.24$ & $+2.5$ [$-0.4$, $+5.4$] \\
\texttt{fork} removed, seven runs, any correct & $48.3\%$ & $4.23$ & $+0.0$ [$-2.9$, $+2.9$] \\
\midrule
\textbf{Octopus} & $\mathbf{48.3\%}$ & $\mathbf{3.74}$ & --- \\
Octopus, earlier evaluation of the same checkpoint & $49.6\%$ & --- & --- \\
\bottomrule
\end{tabular*}
\end{table}

\begin{figure}[!htb]
\centering
\safefig[width=1.0\textwidth]{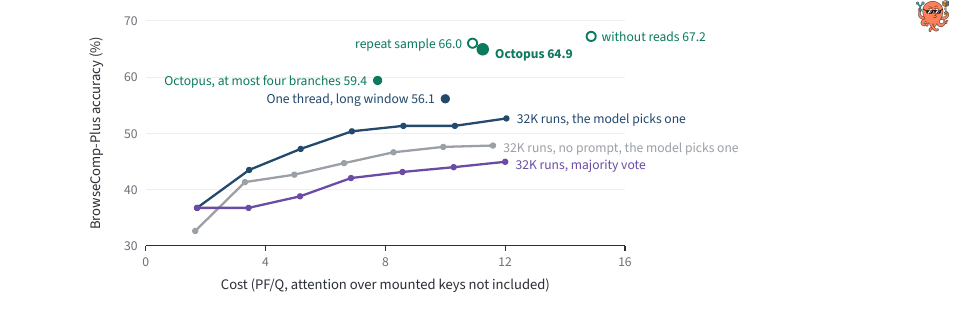}
\caption{Qwen3.6-27B on BrowseComp-Plus with no training: accuracy against cost. Each curve takes one to seven runs of a single $32\text{K}$ thread and chooses one answer.}
\label{fig:zeroshot-full}
\end{figure}

\paragraph{The 27B model in full.}
\Cref{tab:zeroshot-full} and \cref{fig:zeroshot-full} give every agent we ran with Qwen3.6-27B, with the same retrieval, grader and accounting as for the 4B model. Octopus reaches $64.9\%$, $66.0\%$ on a repeat sample, and $59.4\%$ when limited to four branches. Its branches run $35.9$ searches per question against $22.2$ for the thread with a long window. The strategy prompts were tuned on OpenResearcher validation questions, and the single threads have no operation for dropping context, so this comparison shows that the runtime is usable without training and does not isolate the effect of branching. Without reads the same agent scores $67.2\%$, within error of Octopus, at a higher cost: attention over mounted keys adds about $4.3\%$ to the cost of Octopus, and with it included Octopus uses about $21\%$ less compute than the agent without reads.

\begin{table}[!t]
\centering
\caption{Every agent we ran with Qwen3.6-27B on BrowseComp-Plus, with no training. Agents have a strategy prompt unless the row says otherwise. PF/Q does not include attention over mounted keys, and the next column adds it (measured for four branches; $\approx\!11.75$ extrapolates a $4.3\%$ share measured on $200$ questions). The last column is the lead of Octopus in points, paired by question, with its $95\%$ bootstrap interval.}
\label{tab:zeroshot-full}
\setlength{\tabcolsep}{5pt}
\renewcommand{\arraystretch}{1.05}
\begin{tabular*}{\textwidth}{@{\extracolsep{\fill}}lcccc}
\toprule
\textbf{Agent} & \textbf{Accuracy} & \textbf{PF/Q} & \textbf{with mounts} & \textbf{Octopus leads by} \\
\midrule
One thread, $32\text{K}$ window, no prompt & $32.65\%$ & $1.65$ & --- & --- \\
One thread, $32\text{K}$ window & $36.75\%$ & $1.71$ & --- & --- \\
\textbf{Octopus} & $\mathbf{64.94\%}$ & $\mathbf{11.25}$ & $\approx\mathbf{11.75}$ & --- \\
Octopus, repeat sample & $66.02\%$ & $10.91$ & --- & --- \\
Octopus, at most four branches & $59.40\%$ & $7.74$ & $8.21$ & --- \\
Octopus without reads & $67.23\%$ & $14.87$ & $14.87$ & $-2.29$ [$-5.06$, $+0.48$] \\
One thread, long window & $56.14\%$ & $10.00$ & $10.00$ & $+8.80$ [$+5.66$, $+11.81$] \\
Seven $32\text{K}$ runs, the model picks one & $52.65\%$ & $12.04$ & $12.04$ & $+12.29$ [$+9.40$, $+15.18$] \\
Seven $32\text{K}$ runs, majority vote & $44.94\%$ & $12.00$ & $12.00$ & $+20.00$ [$+16.87$, $+23.01$] \\
Seven $32\text{K}$ runs, no prompt, the model picks one & $47.83\%$ & $11.59$ & $11.59$ & $+17.11$ [$+13.86$, $+20.12$] \\
\bottomrule
\end{tabular*}
\end{table}

\section{Cost}
\label{app:compute-curve}

A tree does more work per question than one thread. This appendix reports how accuracy moves with cost, first against the same weights and then against Single-Agent GRPO.

\paragraph{Against the same weights.}
\Cref{tab:curve} varies the branch cap of the trained policy. Each point is compared with runs of the same weights with \texttt{fork} removed, at the smallest number of runs $k^*$ whose cost is at least that of Octopus. With a Bonferroni correction over the three low-cost points fixed before the runs, the lower bounds at caps $4$ and $6$ reach $0.0$. At the full cap of ten, the critical path is $8{,}800$ tokens against $3{,}499$ for seven parallel runs, and seven runs in series take about $1.8\times$ the wall-clock time of Octopus under load. Removing \texttt{rewind} lowers the cost of the trained policy from $3.74$ to $1.66$ PF/Q, and removing reads raises it to $5.56$.

Octopus and repeated runs also reach different questions. Across $829$ scored questions, $327$ are solved by both Octopus and at least one of seven runs, $73$ only by Octopus, $73$ only by the runs, and $356$ by neither, for a union of $57.1\%$.

\begin{table}[!htb]
\centering
\caption{The trained 4B policy on BrowseComp-Plus as the branch cap varies, against $k^*$ runs of the same weights with \texttt{fork} removed, from which either the model or a majority vote picks one answer. Differences are paired by question, with $95\%$ bootstrap intervals.}
\label{tab:curve}
\setlength{\tabcolsep}{5pt}
\renewcommand{\arraystretch}{1.08}
\begin{tabular*}{\textwidth}{@{\extracolsep{\fill}}lcccccc}
\toprule
\textbf{Branch cap} & \textbf{Accuracy} & \textbf{PF/Q} & \textbf{$k^*$} & \textbf{Model picks, $k^*$ runs} & \textbf{Octopus $-$ model picks} & \textbf{Octopus $-$ majority} \\
\midrule
$2$ & $37.7\%$ & $1.50$ & $3$ & $38.0\%$ & $-0.2$ [$-3.0$, $+2.5$] & $+1.1$ [$-1.8$, $+4.0$] \\
$4$ & $42.7\%$ & $2.33$ & $4$ & $39.4\%$ & $+3.3$ [$+0.5$, $+6.0$] & $+4.3$ [$+1.4$, $+7.1$] \\
$6$ & $46.5\%$ & $2.99$ & $5$ & $42.9\%$ & $+3.5$ [$+0.6$, $+6.4$] & $+6.5$ [$+3.6$, $+9.4$] \\
$8$ & $47.6\%$ & $3.50$ & $6$ & $44.3\%$ & $+3.3$ [$+0.5$, $+6.0$] & $+7.1$ [$+4.5$, $+9.9$] \\
$10$ & $48.3\%$ & $3.74$ & $7$ & $45.7\%$ & $+2.5$ [$-0.4$, $+5.4$] & $+7.2$ [$+4.5$, $+10.1$] \\
\bottomrule
\end{tabular*}
\end{table}

\begin{figure}[!htb]
\centering
\safefig[width=1.0\textwidth]{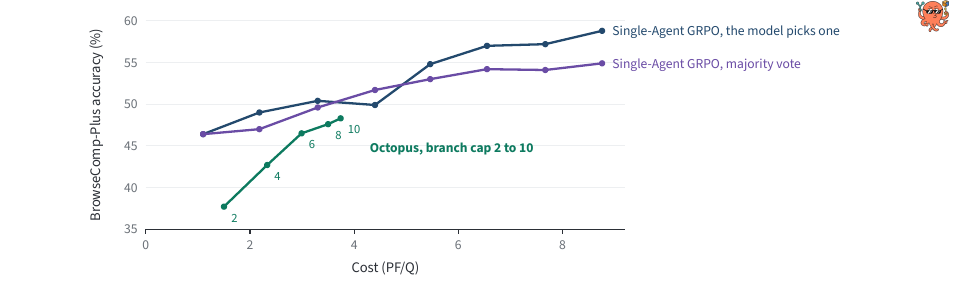}
\caption{Accuracy against cost on BrowseComp-Plus for the trained 4B models: Octopus at branch caps $2$ to $10$, and one to eight runs of Single-Agent GRPO.}
\label{fig:d40-curve}
\end{figure}

\paragraph{Against Single-Agent GRPO.}
\Cref{tab:d40-curve} and \cref{fig:d40-curve} make the same comparison against Single-Agent GRPO. At equal cost, repeated runs of Single-Agent GRPO are ahead at branch caps $2$, $4$ and $6$ and level with Octopus at $8$ and $10$. The $k^*$ rule gives Single-Agent GRPO $10\%$ to $45\%$ more compute than Octopus at every point. With the largest $k$ whose cost is no more than that of Octopus, the differences are $-8.7$ [$-12.1$, $-5.4$] at cap $2$, $-6.4$ [$-9.5$, $-3.4$] at cap $4$, and within error at caps $6$, $8$ and $10$ ($-2.7$, $-2.8$, $-2.1$).

\begin{table}[!htb]
\centering
\caption{Single-Agent GRPO on BrowseComp-Plus by number of runs $k$ (left), and Octopus at each branch cap minus $k^*$ runs of Single-Agent GRPO from which the model picks one answer (right; negative means Single-Agent GRPO is ahead).}
\label{tab:d40-curve}
\setlength{\tabcolsep}{5pt}
\renewcommand{\arraystretch}{1.06}
\begin{tabular*}{\textwidth}{@{\extracolsep{\fill}}cccc@{\hspace{20pt}}cccc}
\toprule
$k$ & \textbf{PF/Q} & \textbf{Majority} & \textbf{Model picks} & \textbf{Branch cap} & $k^*$ & \textbf{PF/Q} & \textbf{Difference} \\
\midrule
$1$ & $1.10$ & $46.4\%$ & $46.4\%$ & $2$ & $2$ & $2.18$ & $-11.3$ [$-14.6$, $-8.2$] \\
$2$ & $2.18$ & $47.0\%$ & $49.0\%$ & $4$ & $3$ & $3.30$ & $-7.7$ [$-10.8$, $-4.5$] \\
$3$ & $3.30$ & $49.6\%$ & $50.4\%$ & $6$ & $3$ & $3.30$ & $-4.0$ [$-7.1$, $-0.8$] \\
$4$ & $4.40$ & $51.7\%$ & $49.9\%$ & $8$ & $4$ & $4.40$ & $-2.3$ [$-5.4$, $+0.8$] \\
$5$ & $5.46$ & $53.0\%$ & $54.8\%$ & $10$ & $4$ & $4.40$ & $-1.6$ [$-4.7$, $+1.6$] \\
$6$ & $6.55$ & $54.2\%$ & $57.0\%$ & & & & \\
$7$ & $7.67$ & $54.1\%$ & $57.2\%$ & & & & \\
$8$ & $8.76$ & $54.9\%$ & $58.8\%$ & & & & \\
\bottomrule
\end{tabular*}
\end{table}

\section{Per-Architecture Integration}
\label{app:arch-blueprints}

\begin{figure}[!htb]
\centering
\safefig[width=1.0\textwidth]{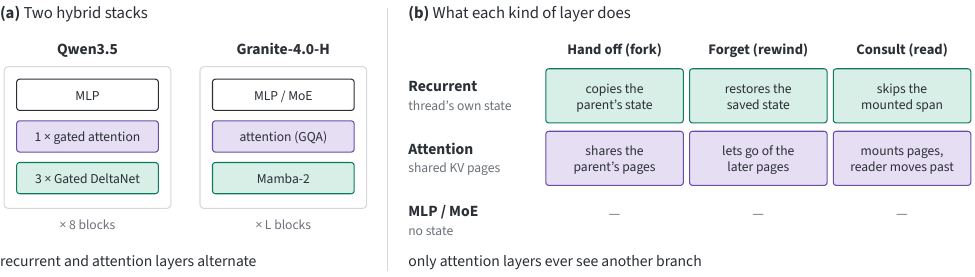}
\caption{The two hybrid families under the three operations. Gated DeltaNet in Qwen3.5 and Mamba-2 in Granite-4.0-H play the same part: they hold the state that belongs to one thread.}
\label{fig:arch-hybrid}
\end{figure}

Octopus supports dense, MoE, and hybrid families through a declarative architecture registry with automatic LoRA target resolution (\cref{tab:arch-hooks}). Pure-attention models have no recurrent layers to keep the reader's task, and mounting into all layers can pull the reader off it, so Octopus mounts into a sparse subset of middle layers ($9/36$ layers in Qwen3-4B-Base) and keeps $96.9\%$ argmax agreement with full-context replay. In Qwen3.5 (eight blocks of three GDN layers and one gated attention layer), all $8$ attention layers mount under the position rule of \cref{eq:k2-pos} and all $24$ GDN layers clone or restore state on \texttt{fork} and \texttt{rewind} and skip mounted spans (\cref{fig:arch-hybrid}). Mamba-2 hybrids follow the same pattern with chunked linear attention ($98.5\%$ token agreement for transitive mounts on Granite-4.0-H).

\begin{table}[!htb]
\centering
\caption{Per-architecture integration hooks in Octopus.}
\label{tab:arch-hooks}
\footnotesize
\setlength{\tabcolsep}{4pt}
\renewcommand{\arraystretch}{1.1}
\resizebox{\textwidth}{!}{%
\begin{tabular}{L{2.7cm} L{3.4cm} L{4.2cm} L{4.2cm}}
\toprule
\textbf{Architecture} & \textbf{Layers} & \textbf{Serving mount and state hook} & \textbf{Prefix-tree training hook} \\
\midrule
\texttt{Qwen3} (Dense) & $L$ Full-Attn, no recurrent state & Sparse middle-layer mount ($9/36$ layers) & Dual variable-length FlashAttn pass \\
\texttt{Qwen3} (MoE) & $L$ Full-Attn + Top-$K$ MoE & Attention-only radix mount + offset; no MoE state & Deduplicated tree tokens in grouped GEMM \\
\texttt{Qwen3.5} (Dense) & $8\times[3\text{ GDN}+1\text{ Attn}]$ & $8$ attention layers mount; $24$ GDN clone/restore and skip & Ancestor conv tokens; chunked GDN recurrence, $16\text{K}$ tiling \\
\texttt{Qwen3.5-MoE/Next} & $24$ GDN + $8$ Attn + MoE & $8$ attention layers mount; $24$ GDN skip & $3$-token conv; level-by-level GDN recurrence; grouped MoE \\
\texttt{Granite-4.0-H} & Mamba-2 + GQA + MoE & Transitive attention mounts ($98.5\%$ agreement); Mamba-2 skips mounts & Conv tokens; Mamba-2 mapped to chunked linear attention \\
\bottomrule
\end{tabular}%
}
\end{table}

\end{document}